\def\year{2020}\relax
\documentclass[letterpaper]{article} 
\usepackage{aaai20}  
\usepackage{times}  
\usepackage{helvet} 
\usepackage{courier}  
\usepackage[hyphens]{url}  
\usepackage{graphicx} 
\usepackage{graphicx}  
\usepackage{booktabs}
\usepackage{arydshln}
\usepackage{subfigure}
\usepackage{amsfonts}
\usepackage{xcolor}

\title{Unsupervised Domain Adaptation on Reading Comprehension}
\author{Yu Cao\textsuperscript{\rm 1}, 
Meng Fang\textsuperscript{\rm 2}, 
Baosheng Yu\textsuperscript{\rm 1}, Joey Tianyi Zhou\textsuperscript{\rm 3} \\
\textsuperscript{\rm 1}UBTECH Sydney AI Center, School of Computer Science, FEIT, The University of Sydney, Australia \\
\textsuperscript{\rm 2} Tencent Robotics X \\ 
\textsuperscript{\rm 3} Institute of High Performance Computing, A*STAR, Singapore \\ 
ycao8647@uni.sydney.edu.au, mfang@tencent.com, baosheng.yu@sydney.edu.au, zhouty@ihpc.a-star.edu.sg 
}
 
\begin{document}

\maketitle

\begin{abstract}
Reading comprehension (RC) has been studied in a variety of datasets with the boosted performance brought by deep neural networks. However, the generalization capability of these models across different domains remains unclear. To alleviate the problem, we investigate unsupervised domain adaptation on RC, wherein a model is trained on the labeled source domain and to be applied to the target domain with only unlabeled samples. We first show that even with the powerful BERT contextual representation, a model can not generalize well from one domain to another. To solve this, we provide a novel conditional adversarial self-training method (CASe). Specifically, our approach leverages a BERT model fine-tuned on the source dataset along with the confidence filtering to generate reliable pseudo-labeled samples in the target domain for self-training. On the other hand, it further reduces domain distribution discrepancy through conditional adversarial learning across domains. Extensive experiments show our approach achieves comparable performance to supervised models on multiple large-scale benchmark datasets.
\end{abstract}

\section{Introduction}
Reading comprehension (RC) is a widely studied topic in Natural Language Processing (NLP) due to its value in human-machine interaction. 
In past relevant research, a variety of large-scale RC datasets were proposed, e.g., \textsc{CNN/DailyMail}~\cite{cnndaily}, \textsc{SQuAD}~\cite{squad}, \textsc{NewsQA}~\cite{newsqa}, \textsc{CoQA}~\cite{coqa} and \textsc{DROP}~\cite{drop}. With a large number of annotations, these datasets make training end-to-end deep neural models possible~\cite{rnet,qanet}. Recent studies also show that BERT~\cite{bert} model achieves higher answer accuracy than human on \textsc{SQuAD}.

However, only unlabeled data is available in many real-world applications. It is a common challenge that machine can learn knowledge well enough in one domain and then answer questions in other domains without any labels. Unfortunately, the generalization capabilities of some existing RC neural models were proven to be weak across different datasets~\cite{yogatama_general}. In fact, the same conclusion can be drawn for BERT according to our experiment, e.g., the performance drops on \textsc{CNN} dataset using the model trained on \textsc{SQuAD}. Therefore, studies to eliminate such performance gaps between various datasets deserve effort.


A potential direction to handle it is transferring knowledge from a labeled source domain to a different unlabeled target domain, which is known as unsupervised domain adaptation~\cite{transfersurvey}, leveraging data from both domains.
However, very few works have studied unsupervised domain adaptation on RC tasks. Although \citeauthor{qa_unsupervised} adapted models using a vanilla self-training, its self-labeling approach cannot ensure the labeling accuracy on the target domain that differs much from the source domain. Besides, it is only applied to some small RC datasets, so its effectiveness on large-scale datasets remains unclear and no general representation is learned. Research on large datasets is more meaningful, since they contains more different patterns than small ones. They pose a greater challenge and better fitting realistic conditions, being the basis to build strong deep neural models. In addition, analyzing the possible influential factors for transfer is also necessary, which provides guide for adaptation. Nevertheless, very limited works contribute to it~\cite{multiqa}.

In this paper, to make use of numerous unlabeled samples in real applications, we focus on unsupervised domain adaptation on large RC datasets.
We propose a novel adaptation method, named as Conditional Adversarial Self-training (CASe).
A fine-tuned BERT model will be obtained on the source domain firstly. Then specifically, in the adaptation stage, an alternated training strategy is applied, containing self-training and conditional adversarial learning in each epoch. The pseudo-labeled samples of the target dataset generated by the last model along with low-confidence filtering will be used for self-training. Compared to the method in~\cite{qa_unsupervised}, the filtering prevent model from learning error target domain distribution especially for large datasets. The conditional adversarial learning, whose discriminator input combines BERT features and final output logits, is utilized because the conditioning generates more comprehensive information than feature only. It encourages the model to learn generalized representations and avoid overfitting on the pseudo-labeled data.

Moreover, we test the generalization of BERT among 6 large RC datasets to prove the importance of adaptation since it fails under most conditions. 
The influential factors that caused the failure are also illustrated via analysis.
We validate the proposed method on different pairs of these 6 datasets, and demonstrate the baseline performance.

Our contributions can be summarized as:
\begin{itemize}
\item We propose a new unsupervised domain adaptation method on RC, which is alternated-staged including self-training with low-confidence filtering and conditional adversarial learning.
\item We experimentally evaluate the method on 6 popular datasets, and it shows a comparable performance to models trained on target datasets, which can be regarded as a pioneer study and a baseline for future work.\footnote{Code available at: \url{https://github.com/caoyu1991/CASe}}
\item We show the transferability among different datasets not only depends on corpora, but also is affected by question forms significantly.
\end{itemize}

\section{Related Work}

Numerous models were proposed for RC tasks. R-NET integrates mutual attention and self-attention into RNN encoder to refine the representation~\cite{rnet}. QANET~\cite{qanet} leverages similar attention in a stacked convolutional encoder to promote performance. BERT~\cite{bert} stacks multiple transformers~\cite{transformer}. By applying unsupervised pre-training tasks and then fine-tuning on specific dataset, it achieves state-of-the-art performance in various NLP tasks including RC. However, none of them explores the model generalizability across different datasets, and their transferabilities still remain unknown.

Prior work on domain adaptation has been done for several NLP tasks.
Some works apply instance weighting on statistical machine translation (SMT)~\cite{mtda_discriminative} or cross-language text classification~\cite{da_textclassification}. Cross-entropy based method is used to select out-domain sentences for training SMT~\cite{mtda_pseudo}. There are also attempts for RC, showing that the performance of RC models on small datasets can be improved by supervised transferring from a large dataset~\cite{qa_transfer,da_qa} using annotations from both domains. MultiQA~\cite{multiqa} strengthens the generalizability of RC model by training on samples from various datasets. Though some studies concentrate on the generalization of RC models and analyze their performance on multiple datasets~\cite{yogatama_general,multitask_nlu}, they do not analyse the influential factors in detail. A parallel work for RC unsupervised domain adaptation~\cite{qa_unsupervised} utilizes a simple self-labeling for re-training, and it is evaluated on 3 small datasets containing thousands of samples.

Many relevant works focus on unsupervised domain adaptation for general CV tasks. Co-training~\cite{co-training} uses two classifiers and two data views to generate labels for unlabeled samples. Both tri-training~\cite{tri-training} and asymmetric tri-training~\cite{asymtri} extend co-training by using three classifiers to generate labels, i.e., labels will be added if two classifiers make an agreement. Some approaches try to learn domain-invariant representations by selecting similar instances between domains or adding a classifier to distinguish domains~\cite{disinvariantfeature,dabackprop}. ADDA~\cite{adda} leverages the Generative Adversarial Networks (GANs) loss on domain label to train a new network. CDAN~\cite{cdan} applies conditional adversarial learning which combines features and labels using a multilinear mapping.

Our work is part of research on unsupervised domain adaptation as well as generalization analysis, with an emphasis on large-scale reading comprehension datasets.

\section{Problem Definition}
We first describe a standard text-span-based RC task such as SQuAD~\cite{squad}. Given a supporting paragraph $\mathcal{P} = \left\langle {p_1},{p_2},...,{p_M}\right\rangle$ with $M$ tokens and a query $\mathcal{Q} = \left\langle {q_1},{q_2},...,{q_L}\right\rangle$ with $L$ tokens, the answer $\mathcal{A} = \left\langle {p_{a^{s}}},{p_{{a^{s}} + 1}},...,{p_{a^{e}}}\right\rangle$ is a text piece in the original paragraph. This task aims to find out the correct answer span $(a^{s},a^{e}), 0 \le a^{s} \le a^{e} \le M$. It means that a model needs to predict two values: the start index and the end index of the answer span.

Unsupervised domain adaptation for RC then is formally defined as follows. There is a source domain with labeled data and a target domain with unlabeled data. We have $n$ labeled samples $\{ (x_i,y_i)\} _{i = 1}^{{n}}$ in the source domain, in which text $x_i=(\mathcal{P}_i, \mathcal{Q}_i)$ and label $y_i=(a^s_i,a^e_i)$, and $n'$ unlabeled target domain samples $\{ (x_j')\} _{j = 1}^{{n'}}$, sharing the same standard RC task as described above. We assume that the data in the source domain is sampled from distribution $\mathcal{D}({x},{y})$ and the data in the target domain is sampled from distribution $\mathcal{D}'({x'},{y'})$, $\mathcal{D} \ne \mathcal{D}'$. Our goal is to find a deep neural model that can reduce the distribution shift and achieve the optimal performance on the target domain. 

\begin{figure*}[ht]
    \centering
    \includegraphics[width=0.94\textwidth]{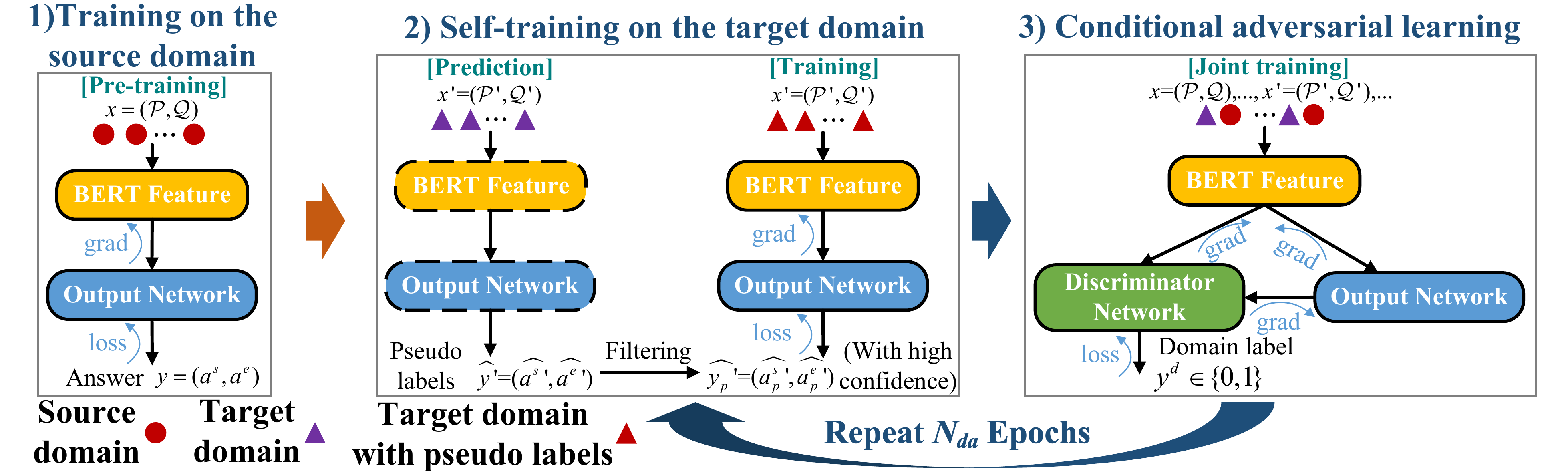}
    \caption{Framework of CASe. (Solid boxes: parameters will be updated. Dashed boxes: parameters will not be updated)}
    \label{fig:framework}
\end{figure*}

\section{Domain Adaptation Method}

The main purpose of our approach is to provide a way to transfer a model trained on labeled data from the source domain to the unlabeled target domain.
Generally, the model with good generalization can reduce the discrepancy of intermediate states generated from different distributions~\cite{da_theory}. We use the BERT model~\cite{bert}, which is a pre-trained contextual model based on unsupervised NLP tasks with a huge 3.3-billion-word corpus. Its model depth and huge training data size ensure that it can generate universal feature representations under a variety of linguistic conditions. And we consider applying adversarial learning to minimize cross-domain discrepancy between $\mathcal{D}({x},{y})$ and $\mathcal{D}'({x'},{y'})$~\cite{adda}. Moreover, pseudo-label based self-training~\cite{self-training} with low-confidence filtering is also utilized for further leveraging unlabeled data in the target domain.

The framework of the proposed Conditional Adversarial Self-training (CASe) approach for unsupervised domain adaptation on RC is illustrated in Figure~\ref{fig:framework}. Our model has three components: a BERT feature network, an output network, and a discriminator network. There are 3 steps in CASe. Firstly, we fine-tune the BERT feature model and output network on the source domain. Secondly, we use self-training on the target domain to get distribution-shifted model. Thirdly, we apply conditional adversarial learning on both domains to further reduce feature distribution divergence. The second and third steps will be proceed iteratively.


\subsection{Training on the Source Domain}
Since we have the labeled data in the source domain, we extend and fine-tune the unsupervised pre-trained base BERT model on these samples.
The BERT feature ${\overline{\bf{f}}} \in {\mathbb{R}^{m \times d}}$ is firstly obtained, in which $m$ and $d$ are the maximum input sequence length and the hidden state dimension in BERT respectively. Then a single-layer linear output network with 2-dimension output vector is added following BERT. One of its output value is used as the answer start logits ${\bf g}^s \in {\mathbb{R}^m}$ and the other one is used as the answer end logits ${\bf g}^e \in {\mathbb{R}^m}$.
Finally, the supervised pre-trained BERT model and output network can be obtained by optimizing the following loss function:
\begin{equation}
\mathcal{L} = \frac{1}{2}\left({f_{CE}({{\bf{g}}^s},{a^s}) + f_{CE}({{\bf{g}}^e},{a^e})}\right),
\end{equation}
where $f_{CE}$ is the cross entropy loss function, $a^s$ and $a^e$ are labels for the answer start and end indices, respectively.

To further enhance the regularization of BERT, we add a batch normalization layer~\cite{batchnorm} between the BERT feature ${\overline{\bf{f}}} \in {\mathbb{R}^{m \times d}}$ and the output network.

\subsection{Self-training on the Target Domain}

After obtaining the pre-trained model from the source domain, we use it to predict sample labels in the target domain. Although data distribution is possibly different between domains, we can still make an assumption that different domains share some similar characteristics. That is, some predicted answers will be similar to or the same as correct answer spans even in a new domain. These predictions combined with corresponding samples $x'=(\mathcal{P}, \mathcal{Q})$ in the target domain, named as pseudo-labeled samples, can be used to teach the model about a new distribution.

Similar to the method in asymmetric tri-training~\cite{asymtri}, to avoid significant error propagation, we select predictions of high confidence as pseudo labels. Since our model generates probabilities for every predicted answer start and end index, a threshold ${T_{prob}}$ will be employed to filter low-confidence samples.

Normally, we apply a softmax function to all output logits and regard generated values as possibilities for indices being the answer start or end index. However, the passage length is usually very large in RC tasks, leading to a very small probability value for each index. This method reduces the numerical distinctions between possibilities and brings more noise, which affects the effectiveness of threshold-based filtering. We thus select a set $\mathcal{U}$ of $n_{best}$ start and end index pairs firstly. These pairs have top-$n_{best}$ sums of start index logits $g_{i}^s$ and end index logits $g_{j}^e,0 \le i \le j \le M$ for corresponding answer spans involved in the target domain, i.e.,
\begin{equation}
    \mathcal{U}=\{ {(i,j)_1,...,(i,j)_{n_{best}}}\} = \mathop {\arg {{\max }_{{n_{best}}}}}\limits_{(i,j)} (g_{{i}}^s + g_{{j}}^e).
\end{equation}
A softmax function then is applied to these $n_{best}$ sums. The span with the highest value after softmax will be regarded as the predicted span and its value is defined as the generating probability $p^{g}$ for current sample, i.e.,
\begin{equation}
    {p^{g}} = \max ({\rm{softmax}}(\{ g_i^s + g_j^e\} )),(i,j) \in \mathcal{U}.
\end{equation}
Samples with $p^{g} \ge T_{prob}$ will be put into pseudo-labeled sample set using the predicted start and end indices as their labels, $\widehat {{a^s}'}$ and $\widehat {{a^e}'}$. The model is trained similar to (1), but $a^s$ and $a^e$ are replaced by $\widehat {{a^s}'}$ and $\widehat {{a^e}'}$, respectively.


In each epoch during adaptation, pseudo-labeled samples are always generated by the last model and previous ones will be abandoned, while ${T_{prob}}$ keeps the same.

\subsection{Conditional Adversarial Learning}\label{adv}

Adversarial learning leverages a discriminator to predict domain classes. But most models only use feature representations for prediction~\cite{adda,dabackprop}, which may be insufficient because the joint distribution of features and labels are not identical across domains. 

\begin{figure}
    \centering
    \includegraphics[width=0.4\textwidth]{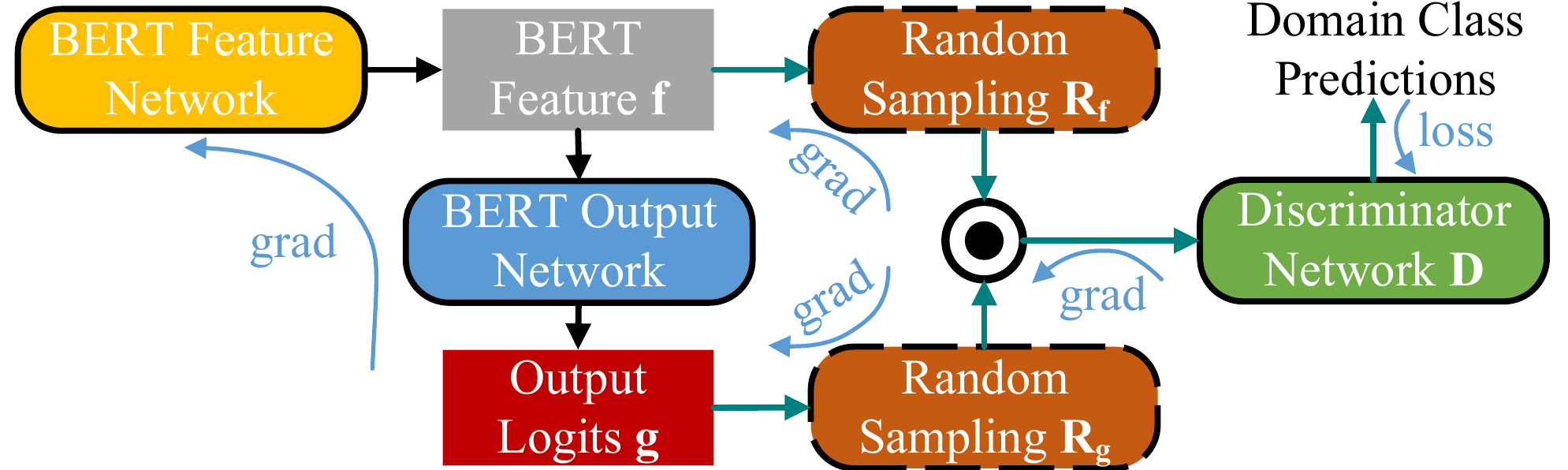}
    \caption{Architecture of the conditional adversarial network used in CASe.}
    \label{fig:discriminator}
\end{figure}

Since our span-based RC tasks can be regarded as a multi-class classification problem and the span properties vary across domains, it poses more challenges for discriminators based only on features. Inspired by the Conditional Adversarial Network (CDAN)~\cite{cdan}, we utilize conditional adversarial learning fusing feature ${\bf{f}}$ and output logits ${\bf{g}}$ for a comprehensive representation, whose network architecture is illustrated in Figure~\ref{fig:discriminator}. It is noted that ${{\bf{f}} \in {\mathbb{R}^{m \times d}}}$ is the BERT feature after the batch normalization layer.

One approach to condition discriminator $D$ on ${\bf{g}}$ is using multilinear map, which is the outer product ${\bf{x}} \otimes {\bf{y}}$ of two vectors and is superior than concatenation~\cite{multilinearmap}. However, it results in dimension explosion and the output dimension is $m \times d \times 2m$ in our application, which is impossible to be embedded. Following CDAN, we tackle it in a randomized approach. The multilinear map of two pairs of features and outputs can be approximated by
\begin{equation}
    \left\langle {{\bf{f}} \otimes {\bf{g}},{\bf{f}}' \otimes {\bf{g}}'} \right\rangle  \approx \left\langle {{Z_R}({\bf{f}},{\bf{g}}),{Z_R}({\bf{f}}',{\bf{g}}')} \right\rangle,
\end{equation}
where $Z_R$ is a randomly sampled multilinear map and generates a vector of dimension $d_R  \ll  m \times d \times 2m$. Given two randomly initialized matrices fixed during training ${\bf{R}}_{\bf{f}} \in \mathbb{R}^{d_R \times m}$ and ${\bf{R}}_{\bf{g}} \in \mathbb{R}^{d_R \times 2m}$, $Z_R$ can be defined as
\begin{equation}
    {Z_R}({\bf{f}},{\bf{g}}) = \frac{1}{{\sqrt {{d_R}} }}\left( {{{\bf{R}}_{\bf{f}}}{av{g_{{\rm{col}}}}}({\bf{f}})} \right) \circ \left( {{{\bf{R}}_{\bf{g}}}{\bf{g}}} \right).
\end{equation}
Here, ${\bf{g}}={\bf{g}}^s \oplus {\bf{g}}^e \in \mathbb{R}^{2m}$. ${av{g_{{\rm{col}}}}}$ means average along columns, transforming the feature matrix into a vector in ${\mathbb{R}^m}$, $\circ$ is element-wise multiplication. 

The discriminator is a 3-layer linear network, whose final layer has a 1-dimension output with sigmoid as the activation function to get a scalar between 0 and 1. And we directly adopt $Z_R({\bf{f}},{\bf{g}})$ as its input for computation efficiency. 

All 3 components, BERT feature network, output network, and discriminator network, are jointly optimized in this stage because discriminator conditions both features and outputs. The loss function is the binary cross entropy loss 
\begin{equation}
    {\mathcal{L}_{adv}} = y^d\log (\widehat y^d) + (1 - y^d)\log (1 - \widehat y^d),
\end{equation}
where $\widehat {y^d}$ is the prediction value from $D$ for domain label, while $y^d \in \{0,1\}$ is the ground truth label, 0 stands for the source domain and 1 for the target domain. Samples $x, x'$ from both domains will be used for joint training.

However, such an optimization imposes equal importance to different samples, while samples that are hard to transfer will pose negative effect on domain adaptation. We quantify the uncertainty of a sample using entropy $E({\bf{p}}) =  - \sum\nolimits_{i = 1}^M ({p_i^s} \log p_i^s + p_i^e\log p_i^e$), to ensure a more effective transfer. $p_i^s$ and $p_i^e$ are probabilities for $i$-th token being the answer start or end index, which can be obtained by applying softmax to whole output logits ${{\bf{g}}^s}$ and ${{\bf{g}}^e}$. We encourage the discriminator to place a higher priority for samples that are easy to transfer. In other words, samples with lower entropy will have higher weights during the conditional adversarial learning (CASe+E). The adversarial loss function can be reformed using the weight $w$ derived from entropy, i.e.,
\begin{equation}
    {\mathcal{L}_{adv-E}} = {w} \cdot {\mathcal{L}_{adv}}, {w}=1 + {e^{ - E({\bf{p}})}}.
\end{equation}

No matter which loss is employed, the conditional adversarial learning makes the feature model and the output model more transferable and generalizable.

\subsection{Algorithm}
The entire procedure of CASe is shown in Algorithm 1. It is noted that no adversarial learning is included in the last epoch of domain adaptation. This aims to make the final model better fit the target domain, because adversarial learning will enhance generalization while affects fitting in specific domains. In step 16 we balance the label number of different domains by removing samples randomly from the larger dataset in merging to avoid unbalanced training.
\begin{table}[htbp]
    \centering
    \begin{tabular}{l}
        \toprule[1.5pt]
        \specialrule{0em}{0pt}{0pt}
        \textbf{Algorithm 1:CASe.} Given a BERT feature network $F$,\\ 
        \specialrule{0em}{-0.5pt}{-0.5pt}
        an output network $G$, and a discriminator $D$. Pre-\\
        \specialrule{0em}{-0.5pt}{-0.5pt}
        training epoch number is $N_{pre}$ and domain adaptation\\
        \specialrule{0em}{-0.5pt}{-0.5pt}
        training epoch number is $N_{da}$\\
        \specialrule{0em}{0pt}{0pt}
        \midrule
        \textbf{Input:} data in the source domain $\mathcal{S}=\{(\mathcal{P}_i,\mathcal{Q}_i,a^s_i,$\\
        $a^e_i)\}_{i = 1}^n$, data in the target domain $\mathcal{S}'=\{(\mathcal{P}_i',\mathcal{Q}_i')\}_{i = 1}^{n'}$. \\
        \textbf{Output:} Optimal model $F$, $G$ in the target domain\\
        1 \ \ \textbf{for} j=1 \textbf{to} $N_{pre}$ \textbf{do}\\
        2 \ \ \quad Train $F$ and $G$ with mini-batch from $\mathcal{S}$ \\
        3 \ \ \textbf{end for} \\
        4 \ \ \textbf{for} j=1 \textbf{to} $N_{da}$ \textbf{do}\\
        5 \ \ \quad Pseudo labeled set $\mathcal{S}^P=\emptyset$ \\
        6 \ \ \quad \textbf{for} k=1 \textbf{to} $n'$ \textbf{do} \\
        7 \ \ \qquad Use $F$, $G$ to predict the label $\widehat {{a_k^s}'}$ and $\widehat {{a_k^e}'}$ for \\
        \ \ \ \ \ \quad \qquad $(\mathcal{P}_k',\mathcal{Q}_k')$ and get probability $p^g_k$\\
        8 \ \ \qquad \textbf{if} $p^g_k \ge T_{prob}$ \textbf{do} \\
        9 \ \ \quad \qquad Put $(\mathcal{P}_k',\mathcal{Q}_k',\widehat {{a_k^s}'},\widehat {{a_k^e}'})$ into $\mathcal{S}^P$ \\
        10 \qquad \textbf{end if} \\
        11 \quad \textbf{end for} \\
        12 \quad \textbf{for} mini-batch $\mathcal{B}$ \textbf{in} $\mathcal{S}^P$ \\
        13 \qquad Train $F$ and $G$ with mini-batch $\mathcal{B}$ \\
        14 \quad \textbf{end for} \\
        15 \quad \textbf{if} j $< N_{da}$ \textbf{do} \\
        16 \qquad $\mathcal{R}=(\{(\mathcal{P}_i,\mathcal{Q}_i)\}_{i = 1}^{n}) \cup \mathcal{S}'$ \\
        17 \qquad \textbf{for} mini-batch $\mathcal{B}$ \textbf{in} $\mathcal{R}$ \\
        18 \quad \qquad Train $F$,$G$,$D$ with $\mathcal{B}$ and domain labels \\
        19 \qquad \textbf{end for} \\
        20 \quad \textbf{end if} \\
        21 \textbf{end for} \\
        \bottomrule[1.5pt]
    \end{tabular}
    \label{tab:my_label}
    \vspace{-0.2cm}
\end{table}

\section{Experiments}
We first evaluate the generalization of BERT among 6 recently released RC datasets and analyze  influential factors. Then we show the performance of proposed CASe for unsupervised domain adaptation on these datasets, along with ablation study and the effects of hyperparameters.

\subsection{Data}

\textsc{\textbf{SQuAD}}~\cite{squad} contains 87k training samples and 11k validation (dev) samples, with questions in natural language given by workers based on paragraphs from Wikipeida. Answers are in text span forms.

\noindent \textsc{\textbf{CNN}} and \textsc{\textbf{DailyMail}}~\cite{cnndaily} contains 374k training  and 4k dev samples, 872k training and 64k dev samples respectively. Their questions are in cloze forms and answers are masked entities in passages.

\noindent \textsc{\textbf{NewsQA}}~\cite{newsqa} contains 120k samples in total, in which QA pairs were generated by crowded workers in natural forms with text spans based on stories from \textbf{CNN}.

\noindent \textsc{\textbf{CoQA}}~\cite{coqa} contains 109k training samples and 8k dev samples, questions are given as conversation forms with multiple turns and answers are in various types including text spans and yes/no.

\noindent \textsc{\textbf{DROP}}~\cite{drop} contains 77k training samples and 9.5k dev samples, given by workers on Wikipedia. It mainly focuses on numerical reasoning and involves answers in numbers or dates except text spans.

Since \textsc{CNN} and \textsc{DailyMail} is much larger than other datasets, we uniformly sampled subsets from two datasets as data source to speed up experiments. The keep ratio is 1/4 and 1/10 respectively, resulting in similar scales as others.

In addition, we pre-processed samples to conduct answer spans for several datasets.
The answers in \textsc{CNN} and \textsc{DailyMail} are mask symbols such as "\textit{@entity1}" which may appears several times in the text. We use a heuristic method to extract spans: 1) find all position indices $\{a_i\}$ of answer masks in a passage; 2) find all position indices $\{\{e^1_i\},...,\{e^K_i\}\}$ of all $K$ question entities in passage; 3) calculate the sum of absolute index distances between an answer appearance $a_j$ and every question entity nearest to it, and $a_j$ with the smallest sum will be used as answer index. 
All masks in these two datasets are also replaced with homologous original tokens.
\textsc{CoQA} contains answers not in text span form. We follow the F1-socre-based method in original paper to obtain the best answer spans. And the concatenation of all previous QA pairs along with the original question in current turn is used as new question. Samples with yes/no as answers or no answer span being found will be discarded. Similarly, we only remain answerable questions with text spans as answers in \textsc{NewsQA} and \textsc{DROP}.

The characterizations of 6 processed datasets are shown in Table~\ref{datasets}. \textsc{DROP} is significantly smaller than others because answers of quantitive reasoning samples are not extractive.

\begin{table}[]
\begin{center}
\begin{tabular}{|l|c|c|c|c|}
\hline \bf Dataset & Train & Dev & Corpus & Question \\
\hline \textsc{SQuAD} & 87,599 & 10,570 & Wikipedia & crowd \\
\textsc{CNN} & 93,627 & 3,833 & CNN news & cloze \\
DailyMail & 87,253 & 6,372 & Daily mail & cloze \\
\textsc{NewsQA} & 76,341 & 4,327 & CNN news & crowd \\
\textsc{CoQA} & 86,077 & 6,272 & Multiple$^*$ & crowd \\
\textsc{DROP} & 28,267 & 3,389 & Wikipedia & crowd \\
\hline
\end{tabular}
\end{center}
\caption{\label{datasets} Characterizations of datasets {\bf after processing}. (*Including corpus from MCTest, CNN, Wikipedia etc.)}
\end{table}

\subsection{Implementation Detail}

We implement CASe based on the BERT implementation in PyTorch by Hugging Face,
using the \textit{base-uncased} pre-trained model with 12 layers and 768-dim hidden state. The maximum input length $m$ is 512 in which the maximum query length is 40. The random sampling dimension $d_R$ is 768. The input dimension of the first layer in the adversarial network is 768. And its intermediate dimension is 512, using ReLU as the activation function in first two layers. Generating probability threshold $T_{prob}$ is set as 0.4 and $n_{best}=20$. Adam optimizer~\cite{adam} is employed with learning rate $3\times10^{-5}$ in the source domain training, $2\times10^{-5}$ in the self-training and $10^{-5}$ in the adversarial learning, with batch size 12. A dropout with rate 0.2 is applied on both the BERT feature network and the discriminator. We set the epoch number $N_{pre}=3$ in pre-training and $N_{da}=4$ in domain adaptation.

Besides, since the input length may be larger than $m$, we truncate a passage using a sliding window to fit the input length whose moving step is 128. And text pieces excluding the answers will be discarded in training.

\subsection{Generalization and Influential Factors}

We firstly test the generalization capability of BERT by fine-tuning it on one dataset and directly applying it to another dataset without any change. We call such models as \textbf{zero-shot} models. The performance on dev sets for transferring among 6 datasets is shown in Table~\ref{generalization}. 

In a high-level observation, the performance of zero-shot models drops significantly in most cases except the transferring between \textsc{CNN} and \textsc{DailyMail}. The average 55.8\% reduction in exact match (EM) and 50.0\% reduction in F1 compared to models trained on the target dataset (\textsc{Self}) prove that BERT cannot generalize well to unseen datasets, despite a huge corpus is used in unsupervised pre-training.

\begin{table*}[t!]
\begin{center}
\begin{tabular}{|l|cccccc|}
    \hline
    Datasets & \textsc{SQuAD} & \textsc{CNN} & \textsc{DailyMail} & \textsc{NewsQA} & \textsc{CoQA} & \textsc{DROP} \\
    \hline
    \textsc{SQuAD} & - & 16.72\,/\,26.42 & 21.12\,/\,21.70 & \bf 40.03\,/\,57.42 & \bf 29.58\,/\,39.58 & \bf 19.06\,/\,29.73 \\
    \textsc{CNN} & 18.97\,/\,24.34 & - & \bf 81.53\,/\,83.59 & 9.38\,/\,15.36 & 7.10\,/\,10.26 & 4.40\,/\,7.50 \\
    \textsc{DailyMail} & 9.72\,/\,14.76 & \bf 77.22\,/\,79.73 & - & 5.89\,/\,10.69 & 5.68\,/\,8.75 & 4.69\,/\,8.02 \\
    \textsc{NewsQA} & 64.80\,/\,78.32 & 25.10\,/\,34.66 & 28.41\,/\,38.44 & - & 27.14\,/\,38.75 & 12.36\,/\,21.00\\
    \textsc{CoQA} & \bf 65.25\,/\,74.92 & 18.21\,/\,24.76 & 22.65\,/\,28.12 & 37.74\,/\,53.85 & - & 14.75\,/\,21.60\\
    \textsc{DROP} & 55.53\,/\,68.36 & 14.32\,/\,22.26 & 17.44\,/\,25.78 & 28.36\,/\,44.35 & 16.15\,/\,24.82 & -\\
    \hline
    \textsc{Self} & 79.85\,/\,87.46 & 82.76\,/\,84.73 & 81.37/\,/\,83.33 & 52.05\,/\,67.41 & 48.98\,/\,63.99 & 44.67\,/\,52.51\\
    \hline
\end{tabular}
\end{center}
\caption{\label{generalization} Performance of zero-shot models on dev set when transferring among datasets. Rows correspond to source datasets and columns to target datasets. \textsc{Self} means training and testing on the same dataset. Left value in each cell is for {\bf exact match (EM)} while the right one is for \bf{F1 score}.}
\end{table*}

\begin{table*}[t!]
\begin{center}
\begin{tabular}{|l|cccccc|}
    \hline
    Datasets & \textsc{SQuAD} & \textsc{CNN} & \textsc{DailyMail} & \textsc{NewsQA} & \textsc{CoQA} & \textsc{DROP} \\
    \hline
    \textsc{SQuAD} & - & \bf 80.64\,/\,82.24 & 80.78\,/\,82.77 & \bf 52.69\,/\,68.15 & \bf 52.38\,/\,67.56 & \bf 50.34\,/\,57.53 \\
    \textsc{CNN} & 79.86\,/\,87.65 & - & \bf 84.26\,/\,86.01 & 48.37\,/\,63.47 & 51.71\,/\,67.09 & 45.59\,/\,53.57 \\
    \textsc{DailyMail} & 79.04\,/\,87.07 & 78.06\,/\,80.36 & - & 50.13\,/\,65.90 & 50.06\,/\,65.76 & 41.69\,/\,50.07 \\
    \textsc{NewsQA} & \bf 80.17\,/\,88.14 & 79.60\,/\,81.57 & 80.93\,/\,82.99 & - & 50.05\,/\,66.49 & 47.36 / 56.42\\
    \textsc{CoQA} & 78.38\,/\,85.93 & 74.75\,/\,76.65 & 76.87\,/\,78.88 & 51.21\,/\,65.83 & - & 42.08\,/\,50.07\\
    \textsc{DROP} & 74.03\,/\,83.35 & 77.09\,/\,79.03 & 80.34\,/\,82.49 & 51.91\,/\,66.95 & 48.90\,/\,64.29 & -\\
    \hline
    \textsc{SQuAD} & - & 80.20\,/\,81.93 & 79.91\,/\,82.06 & \bf 51.56\,/\,66.79 & \bf 50.77\,/\,65.94 & \bf 48.45\,/\,57.33 \\
    \textsc{CNN} & 78.59\,/\,86.39 & - & \bf 83.40\,/\,85.06 & 48.95\,/\,64.45 & 49.38\,/\,64.57 & 44.15\,/\,51.87 \\
    \textsc{DailyMail} & 78.07\,/\,86.22 & \bf 82.44\,/\,84.36 & - & 50.91\,/\,65.90 & 48.64\,/\,63.80 & 41.58\,/\,47.74 \\
    \textsc{NewsQA} & \bf 78.87\,/\,87.06 & 80.49\,/\,82.43 & 80.93\,/\,82.99 & - & 48.01\,/\,64.30 & 45.06\,/\,54.34\\
    \textsc{CoQA} & 78.24\,/\,85.80 & 76.34\,/\,78.22 & 78.12\,/\,79.88 & 50.80\,/\,65.55 & - & 41.43\,/\,49.40\\
    \textsc{DROP} & 74.81\,/\,83.67 & 80.38\,/\,82.21 & 80.78\,/\,82.96 & 50.01\,/\,65.16 & 46.27\,/\,62.67 & -\\
    \hline
    \textsc{Self} & 79.85\,/\,87.46 & 82.76\,/\,84.73 & 81.37/\,/\,83.33 & 52.05\,/\,67.41 & 48.98\,/\,63.99 & 44.67\,/\,52.51\\
    \hline
\end{tabular}
\end{center}
\caption{\label{domain_adaptation} Domain adaptation performance of CASe on dev sets of datasets. The top of the table shows results for CASe+E (Entropy-weighted loss), while the bottom for standard CASe. Rows are source datasets and columns are target datasets. The left value in each cell is \textbf{exact match(EM)}, while right one is \textbf{F1 score}. \textsc{Self} stands for training and testing on the same dataset.}
\end{table*}

Taking a closer look, we can find the reductions vary across different dataset pairs. The drops of transferring among 4 datasets, \textsc{SQuAD}, \textsc{NewsQA}, \textsc{CoQA} and \textsc{DROP}, are smaller than transferring to/from rest 2 datasets, especially from latter 3 ones to \textsc{SQuAD}. And the transferring between \textsc{CNN} and \textsc{DailyMail} achieves equivalent performance to \textsc{Self}. \textsc{CNN} and \textsc{NewsQA} share the same corpus but the transferring fails due to different question forms(natural vs. cloze), and the corpus discrepancy of \textsc{SQuAD} and \textsc{NewsQA} leads to homologous result. On the other hand, the same question forms and similar corpora of \textsc{CNN} and \textsc{DailyMail} make successful transferring. Therefore, it can be concluded that not only the corpus but also the question form affect the generalization. It is also observed that the different focus as well as reasoning types affect the transfer between datasets even with same corpus and question type, i.e. simple single-sentence reasoning in \textsc{SQuAD} vs. complex reasoning (comparison, selection) in \textsc{DROP}. 

\begin{figure}[htbp]
    \centering
    \includegraphics[width=0.35\textwidth]{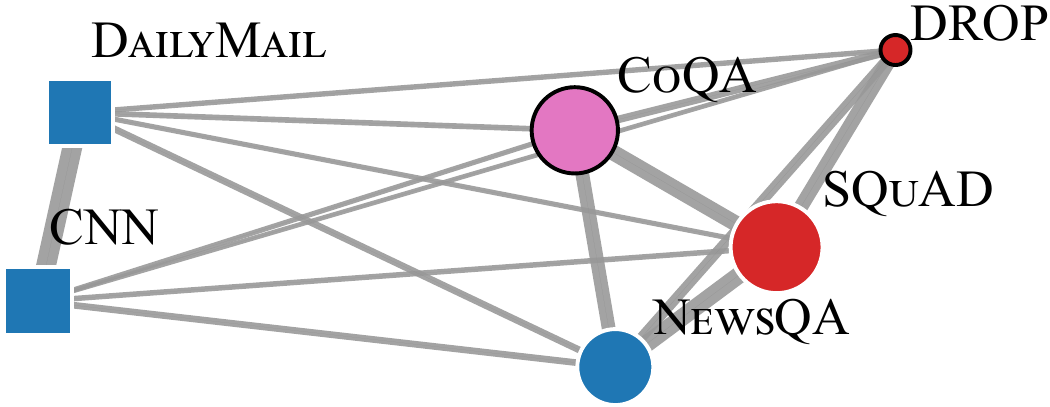}
    \caption{Visualization of relations between datasets based on performance.  Node shape represents question form (rectangle:cloze, circle:natural). Node color represents corpus (red: Wikipedia, blue: news, purple: multiple).}
    \label{fig:visualization}
\end{figure}

We visualize the relations between 6 datasets using force-directed graph in Figure~\ref{fig:visualization}. The force between every two datasets can be calculate via ${F_{ij}} = {{{P_{ij}}} \mathord{\left/{\vphantom {{{P_{ij}}} {{P_j} + }}} \right.\kern-\nulldelimiterspace} {{P_j} + }}{{{P_{ji}}} \mathord{\left/{\vphantom {{{P_{ji}}} {{P_i}}}} \right.\kern-\nulldelimiterspace} {{P_i}}}$. $P_{ij}$ is the average performance of EM and F1 from source dataset $i$ to target dataset $j$, and $P_{i}$ is the average performance of \textsc{Self} model on dataset $i$. Edge widths are positively correlated to force $F$ between nodes, while the size of each node reflects dataset scale. It is noted that datasets cluster more significantly according to {\bf question forms} (node shapes), comparing to {\bf corpora} (node colors) who also affect it.

\subsection{Domain Adaptation Results}
We now evaluate the performance of proposed CASe method for unsupervised domain adaptation on RC datasets, including standard CASe and CASe with entropy-weighted loss in adversarial learning (CASe+E). The results are shown in Table~\ref{domain_adaptation}. Generally speaking, no matter which loss function is used in adversarial learning, CASe achieves significant performance improvement compared to zero-shot models. Despite annotated data is unavailable in the target domain, most results are comparable to \textsc{Self} models, and some of them are even better. In conclusion, CASe transfers knowledge from one domain to another one successfully.

\begin{figure*}[!ht]
    \centering
    \subfigure[Performance varies with $T_{prob}$ (Upper: EM, lower: F1).]{
        \includegraphics[width=0.28\textwidth]{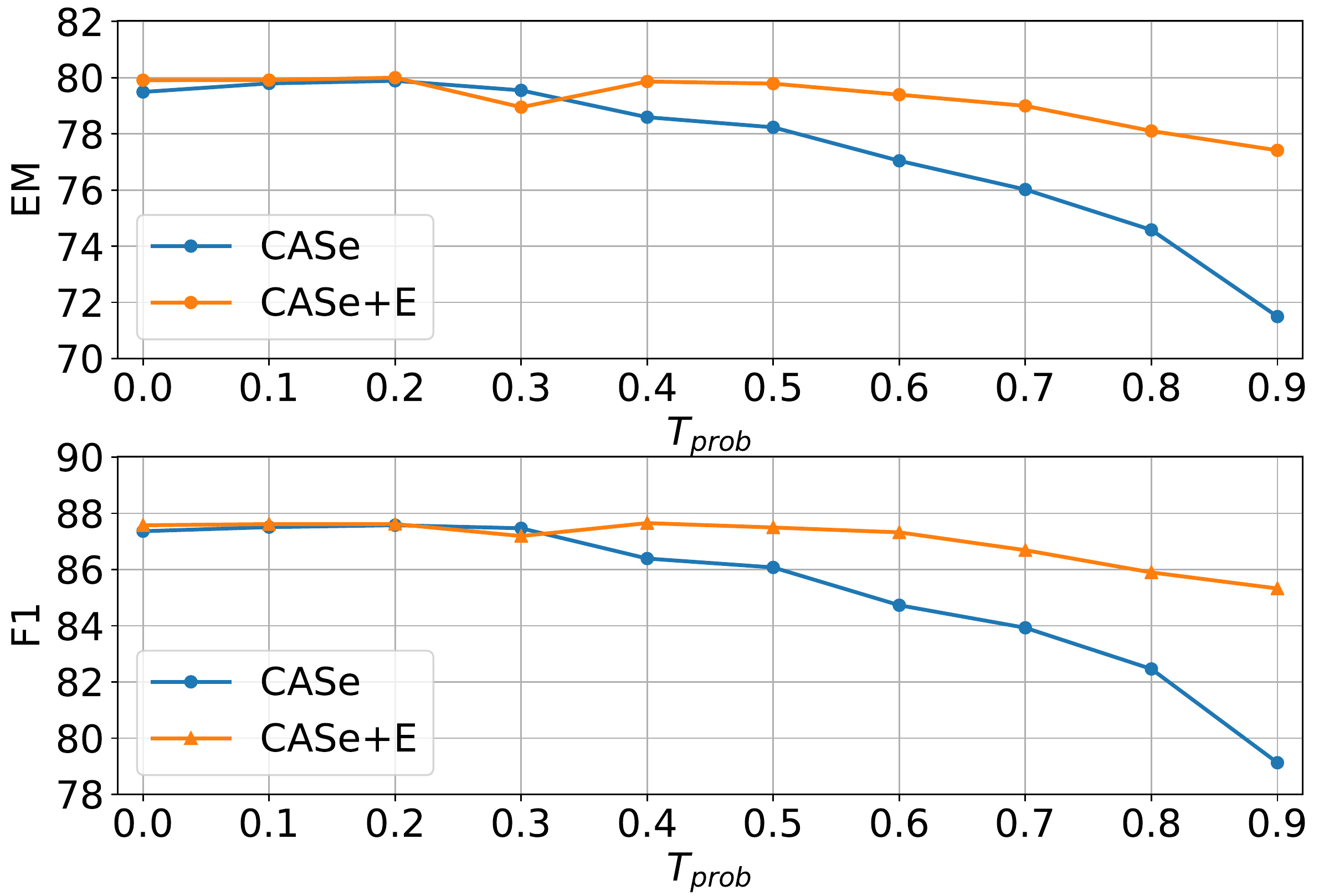}
        \label{fig:performance_tprob}
    }\quad
    \subfigure[Numbers of pseudo-labeled samples generated in each epoch under different $T_{prob}$.]{
        \includegraphics[width=0.32\textwidth]{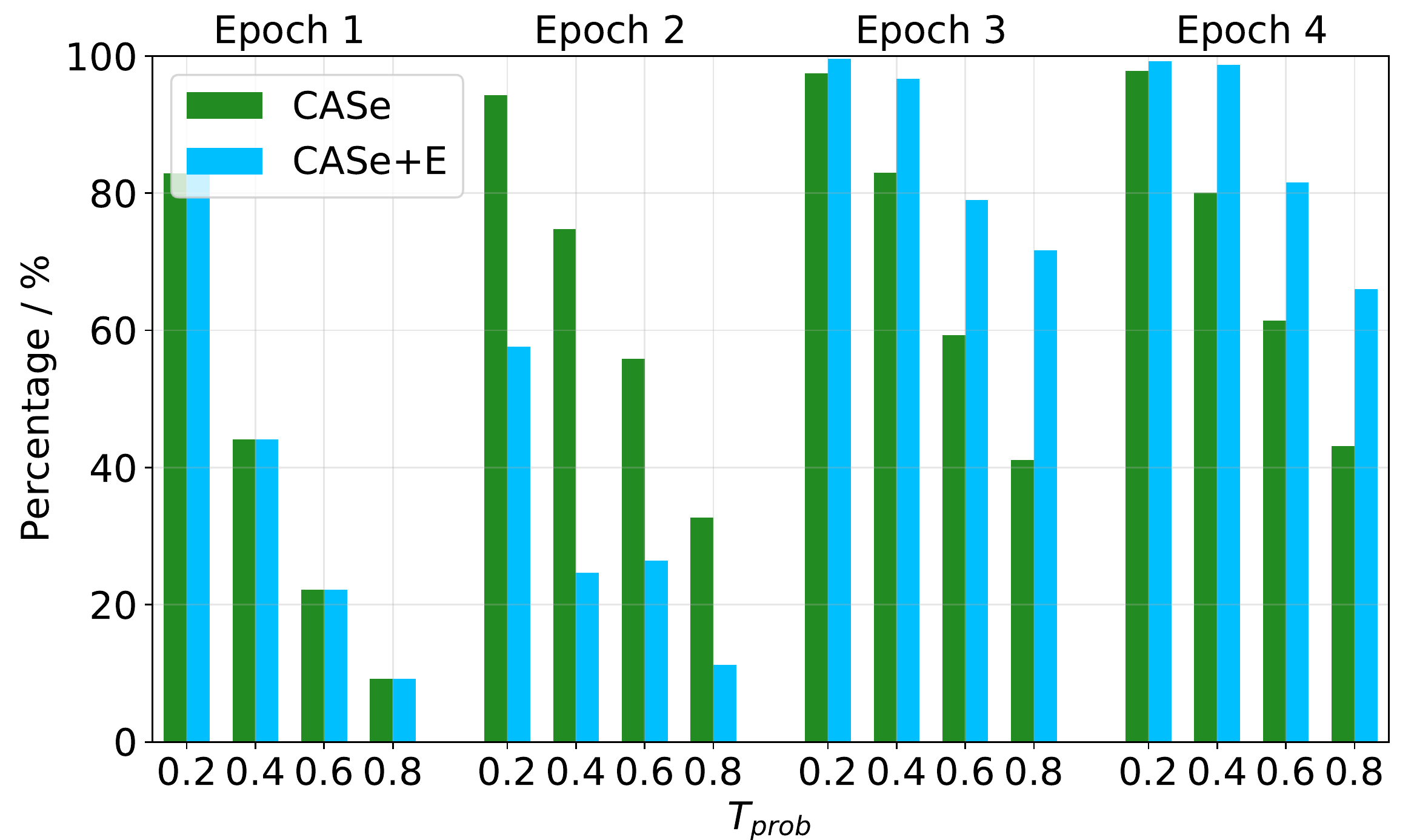}
        \label{fig:number_tprob}
    }\quad
    \subfigure[Performance varies with adaptation stages and epoch numbers when $T_{prob}=0.4$.]{
        \includegraphics[width=0.3\textwidth]{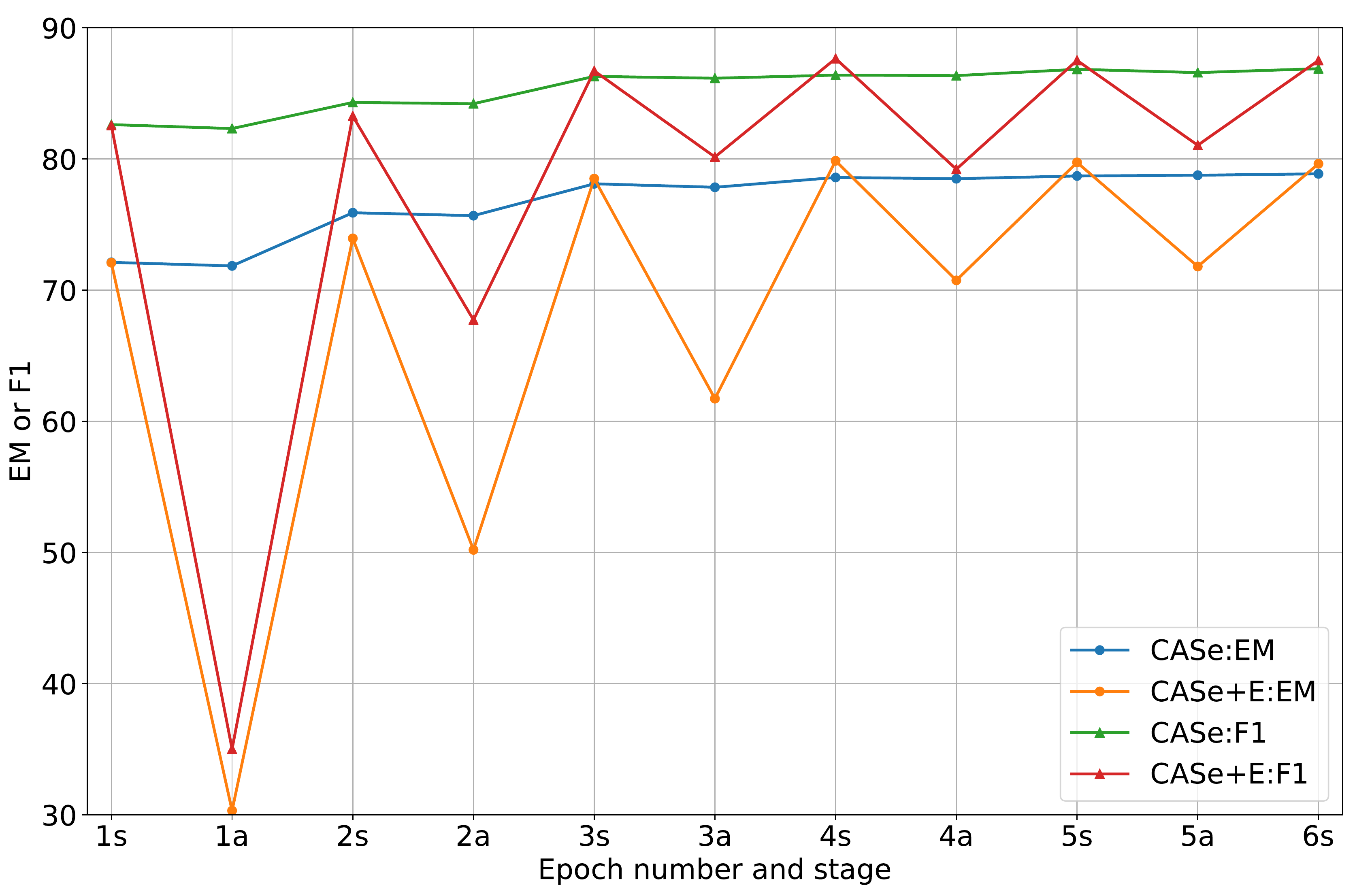}
        \label{fig:performance_epoch}
    }
    \caption{Influence of hyperparameters on adaptation performance of CASe and CASe+E under CNN to SQuAD (\textbf{C$\rightarrow$S}).}
\end{figure*}

Domain adapted models between two very alike datasets, \textsc{CNN} and \textsc{DailyMail}, show a higher EM than \textsc{Self}. They are similar on both corpora and question forms, which means more valid data can be utilized for self-training to get a model with deeper comprehension. Zero-shot model performs poorly when transferring between natural-question-based datasets and cloze-question-based datasets, e.g., \textsc{SQuAD} to \textsc{CNN}. But CASe can nearly eliminate such gaps between transferred model and \textsc{Self} models due to the new distribution  learned in self-training and generalized representation optimized in adversarial learning. The performance of most adaptations on \textsc{CoQA} and \textsc{DROP} is better than \textsc{Self} because they benefit from more extra data.

Entropy-based loss weighting also show its effectiveness because it makes learning focus on samples simple to be transferred so as to obtain more correct knowledge in the target domain. And CASe+E shows 0.5\% to 2\% higher in EM accuracy than CASe under most conditions except some specific dataset pairs such as \textsc{DailyMail} to \textsc{CNN}.

\subsubsection{Ablation Study}
We do ablation test on 4 domain adaptation dataset pairs, which are \textsc{CNN} to \textsc{SQuAD} (\textbf{C$\rightarrow$S}), \textsc{DailyMail} to \textsc{CNN} (\textbf{D$\rightarrow$C}), \textsc{CNN} to \textsc{NewsQA} (\textbf{C$\rightarrow$N}) and \textsc{SQuAD} to \textsc{CoQA} (\textbf{S$\rightarrow$Co}), including adaptation between datasets with same/different question forms and/or corpora. The EM results on ablated models are shown in Table~\ref{ablation}, in which - \textit{conditional} means using unconditional adversarial learning instead of conditional one, while - \textit{Adv learning} for removing whole adversarial learning, - \textit{Self-training} for removing self-training and - \textit{Batch norm} for removing batch normalization, all based on CASe. It is observed that self-training plays the most important role under all configurations. Performance drops without discriminator conditioning on output or whole adversarial learning. Batch normalization has slight effect, removing it promotes the results under two configurations while it has opposite effect under others.

\subsubsection{Generalization after Domain Adaptation}
We test the performance of transferred models on the source datasets to check their generalization, which is shown in Table~\ref{after_adaptation}. 4 datasets pairs in ablation study is involved plus \textsc{NewsQA} to \textsc{DROP} (\textbf{N$\rightarrow$Dr}). There are performance declines compared to models trained on the source datasets, except \textbf{D$\rightarrow$C} in which datasets have very similar properties. It means our CASe method results in a good transferred model at the meantime leads to knowledge loss in the source domain.

\subsubsection{Impact of $T_{prob}$}
Figure~\ref{fig:performance_tprob} demonstrates the performance of CASe and CASe+E on \textbf{C$\rightarrow$S} varied with different generating probability $T_{prob}$ in terms of EM and F1 scores. CASe+E shows higher stability and performance than CASe under different $T_{prob}$. CASe and CASe+E reach their peaks at 0.3 and 0.4 respectively, while both of them show descending trends when $T_{prob} \ge 0.4$. 

The numbers of generated pseudo-labeled samples in every epoch on \textbf{C$\rightarrow$S} with different $T_{prob}$ are shown in Figure~\ref{fig:number_tprob}. Obviously, a lower threshold results in more samples and longer training time. Although CASe generate more samples stably than previous epoch, samples generated by CASe+E may decrease in the 2nd epoch, but more samples will be generated latter compared to CASe. Thus CASe+E achieves better results under most conditions because more valid samples are utilized. Considering the overall performance as well as the trade-off between EM accuracy and complexity, we set $T_{prob}$ as 0.4 in our experiment.

\begin{table}[t]
    \centering
    \begin{tabular}{|l|c|c|c|c|}
         \hline
         & \textbf{C$\rightarrow$S} & \textbf{D$\rightarrow$C} & \textbf{C$\rightarrow$N} & \textbf{S$\rightarrow$Co} \\
         \hline
         CASe+E &  66.46 & 78.06 & 48.37 & 52.38\\
         CASe & 65.24 & 82.44 & 48.95 & 50.77\\
         \quad- \textit{conditional} & 64.47 & 82.26 & 47.31 & 50.25\\
         \quad- \textit{Adv learning} & 65.05 & 81.21 & 47.89 & 49.05\\
         \quad- \textit{Self-training} & 16.55 & 77.07 & 14.26 & 23.81\\
         \quad- \textit{Batch norm} & 65.97 & 81.91 & 48.27 & 51.08\\
         \hline
    \end{tabular}
    \caption{EM results of CASe ablation test on 4 dataset pairs.}
    \label{ablation}
\end{table}

\begin{table}[t]
    \centering
    \begin{tabular}{|l|c|c|c|c|c|}
         \hline
         & \textbf{C$\rightarrow$S} & \textbf{D$\rightarrow$C} & \textbf{C$\rightarrow$N} & \textbf{S$\rightarrow$Co} & \textbf{N$\rightarrow$Dr} \\
         \hline
         CASe+E &  66.37 & 82.19 & 64.65 & 52.97 & 40.07\\
         CASe & 68.61 & 81.61 & 65.43 & 51.48 & 40.17\\
         \hline
         \textsc{Self} & 80.77 & 80.85 & 80.77 & 66.51 & 52.05\\
         \hline
    \end{tabular}
    \caption{EM results on source datasets after adaptaiton.}
    \label{after_adaptation}
\end{table}

\subsubsection{Impact of Epoch Number}
In Figure~\ref{fig:performance_epoch}, we present the performance of CASe and CASe+E after different stages in every epoch on \textbf{C$\rightarrow$S}. E.g., \textit{1s} means result after the self-training stage in 1st epoch, \textit{2a} means results after conditional adversarial learning stage in 2nd epoch. CASe+E shows obvious fluctuations between the self-training and the adversarial learning compared to CASe. Not matter CASe or CASe+E, the performance tends to be saturated after 3 complete epochs. That is the reason why we set $N_{da}$ as 4.

\section{Conclusion}
In this paper, we explore the possibility of transferring knowledge for unsupervised domain adaptation on Reading Comprehension.
Our experiment proves that even the BERT model cannot generalize well between different domains, and the divergence of both corpora and question forms results in this failure.
Then we propose a new unsupervised domain adaptation method, Conditional Adversarial Self-training (CASe). After fine-tuning a BERT model on labeled data from the source domain, it uses self-training and conditional adversarial learning alternately in every epoch to make the model better fit the target domain and reduce the domain distribution discrepancy.
The experimental results among 6 RC datasets demonstrate the effectiveness of CASe. It promotes performance remarkably over zero-shot models, showing similar performance to supervised trained on the target domain.

\section*{Acknowledgements}
We thank Boqing Gong and the anonymous reviewers for insightful comments and feedback.

\fontsize{9.5pt}{10.5pt} 
\selectfont
\bibliographystyle{aaai}
\bibliography{AAAI-CaoY.3066.bib}

\end{document}